\documentclass{article}

 \usepackage[preprint]{neurips_2020}

\usepackage{multirow}
\usepackage[utf8]{inputenc} 
\usepackage[T1]{fontenc}    
\usepackage{hyperref}       
\usepackage{url}            
\usepackage{booktabs}       
\usepackage{amsfonts}       
\usepackage{nicefrac}       
\usepackage{microtype}      
\usepackage{graphicx}
\usepackage{algorithm}
\usepackage{algpseudocode}
\usepackage{subfigure}
\title{Federated Mutual Learning}

\author{
  Tao Shen\\
  Department of Computer Science\\
  Zhejiang University\\
  \texttt{tao.shen@zju.edu.cn} \\
  \And
  Jie Zhang\\
  School of Software Technology\\
  Zhejiang University\\
  \texttt{zjujiezhang@gmail.com} \\
  \And
  Xinkang Jia\\
  School of Software Technology\\
  Zhejiang University\\
  \texttt{Xk\_jia@zju.edu.cn} \\
  \And
  Fengda Zhang\\
  Department of Computer Science\\
  Zhejiang University\\
  \texttt{fdzhang@zju.edu.cn} \\
  \And
  Gang Huang\\
  Artificial Intelligence Research Center\\
  Zhejiang Lab\\
  \texttt{huanggang@zju.edu.cn} \\
  \And
  Pan Zhou\\
  School of Electronic Information \& Communications\\
  Huazhong University of Science and Technology\\
  \texttt{panzhou@hust.edu.cn} \\
  \And
  Kun Kuang\\
  Department of Computer Science\\
  Zhejiang University\\
  \texttt{kunkuang@zju.edu.cn} \\
  \And
  Fei Wu\\
  Department of Computer Science\\
  Zhejiang University\\
  \texttt{wufei@zju.edu.cn} \\
  \And
  Chao Wu\thanks{Corresponding Author}\\
  School of Public Affairs\\
  Zhejiang University\\
  \texttt{chao.wu@zju.edu.cn} \\
}

\begin{document}
\maketitle

\begin{abstract}
Federated learning (FL) enables collaboratively training deep learning models on decentralized data. However, there are three types of heterogeneities in FL setting bringing about distinctive challenges to the canonical federated learning algorithm (FedAvg). First, due to the Non-IIDness of data, the global shared model may perform worse than local models that solely trained on their private data; Second, the objective of center server and clients may be different, where center server seeks for a generalized model whereas client pursue a personalized model, and clients may run different tasks; Third, clients may need to design their customized model for various scenes and tasks; In this work, we present a novel federated learning paradigm, named Federated Mutual Leaning (FML), dealing with the three heterogeneities. FML allows clients training a generalized model collaboratively and a personalized model independently, and designing their private customized models. Thus, the Non-IIDness of data is no longer a bug but a feature that clients can be personally served better. The experiments show that FML can achieve better performance than alternatives in typical FL setting, and clients can be benefited from FML with different models and tasks.
\end{abstract}

\section{Introduction}
The protection of data privacy is increasingly critical in the big data era. It is not only a public concern, but a rule enforced by laws such as \emph{General Data Protection Regulation (GDPR)} in the European Union. Thus, data massively generated in devices (e.g. mobile phones, wearables, IoTs) or in organizations (e.g. hospitals, companies, courts) cannot be gathered in central server, giving rise to a huge challenge for deep learning. Federated learning (FL) \citep{mcmahan2016communication} is a deep learning setting where clients can collaboratively train a shared model under the orchestration of central server, while keeping the data decentralized \citep{kairouz2019advances,lim2020federated,yang2019federated}. It is an emerging technique that helps us out of the dilemma of ``data island'', and gets extensive applications like mobile apps, autopilots, healthcares, and financial services etc. However, researches on federated learning are perplexed by some distinctive challenges, among which we focus on in this paper is the heterogeneity problem. Here we summarize them as three heterogeneities: data, objective, and model (DOM), shown in Figure \ref{dom}:
\begin{figure}[h]
	\centering
	\includegraphics[width=\textwidth]{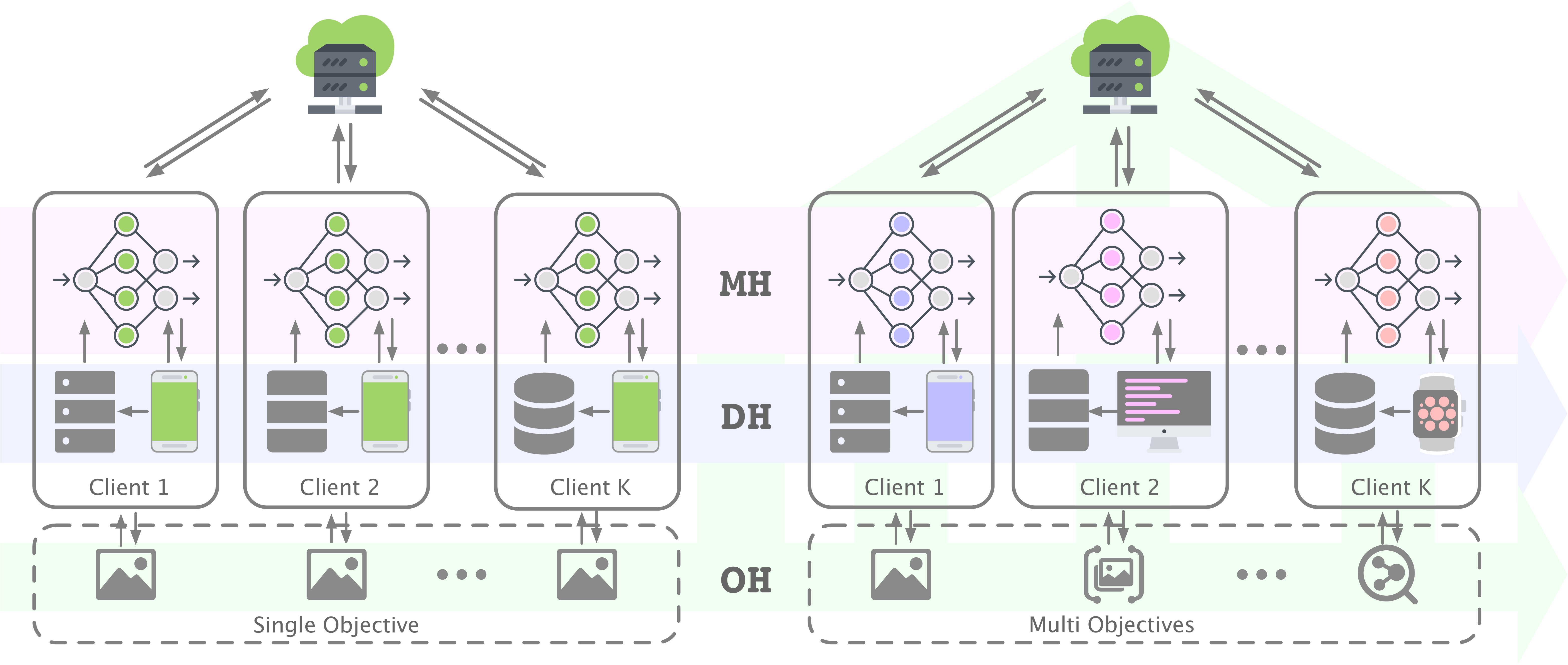}
	\caption{The Non-IIDness of data is usually mentioned in typical federated learning, which refers to the data heterogeneity (DH). Besides, server and clients in FL used to train a single model with same architecture on a single task. However, the objective of server may be different with that of clients and clients may run different tasks (OH). Thus, clients may want to train different models (MH).}\label{dom}
\end{figure}
\begin{itemize}
	\item \textbf{Data Heterogeneity (DH)}: Instead of the independent and identically distributed (\emph{IID}) data in centralized deep learning task, the isolated data in a federated setting are in a \emph{Non-IID} manner. It means that the data, $\{(\mathcal{X}_1, \mathcal{Y}_1), \cdots, (\mathcal{X}_k, \mathcal{Y}_k)\}$, distributed in different clients $i, j \in \{k\}$ may generated by distinct distribution $(x, y) \sim \mathcal{P}_k(x, y)\not= \mathcal{P}_j$. This statistical heterogeneity of data leads to significant accuracy reduction compared to IID data manner, explained by the weight divergence \citep{zhao2018federated} at the stage of model weights averaging. 
	\item \textbf{Objective Heterogeneity (OH)}: The objective of federated learning is ambiguous, where global model and local model serves different objectives. 1) For server and clients: on the one hand, server aims to train a single generalized model (fits joint distribution $\mathcal{P}_{joint}(x, y)$) for all clients and new participants; on the other hand, clients aim to train a personalized model (fits distinct distribution $\mathcal{P}_{k}(x, y)$) for themself. FedAvg \citep{mcmahan2016communication} will compromise the personalities of clients on achieving consensus because of the DH. 2) For clients: each client may have similar features (e.g. visual features) but different tasks (e.g. 10- or 100-classification), but they all want to be benefited from FL but FedAvg is unable to deal with it. 
	\item \textbf{Model Heterogeneity (MH)}: FedAvg, where the global model is aggregated by averaging weights of local models $\sum_{i=1}^{n} \frac{n_k}{n}w^{k}$, cannot fulfill the requirement of customized models for various scenes and tasks. Clients may have different hardware capabilities \citep{wu2019fbnet, he2020fednas}, different representation of local data \citep{liang2020think,gao2019hhhfl}, or different tasks \citep{smith2017federated}, for which they require to design their own models. Besides, local model also involves privacy issues, since it is a kind of private property that should be protected from being stolen. 
\end{itemize}
 
In this work, we propose a novel paradigm for federated learning, named Federated Mutual Leaning (FML), in response to the three heterogeneities (DOM). First, FML deals with data and objective heterogeneities by enabling each client training a personalized models. In the sense of OH, DH is harmful for server but is beneficial for clients, that means the Non-IIDness of data is no longer a bug but a feature that clients can be personally served better; Second, the local personalized models can be benefited from collaboratively training in FML with similar but different tasks. Third, FML allows clients designing their own customized models for various scenes and tasks. The experiments compare FML with other FL methods in typical FL settings and validate the performance of FML in DOM. The results show that FML can outperform other FL methods, and is capable to deal with DOM.

\section{Related Work}
\paragraph{Data Heterogeneity} 
One of the most important difference between federated learning and distributed learning (usually refers to distributed training in data center) is whether the data of clients are fixed locally and cannot be accessed by others. This feature bring the safety of data privacy, but leads to the Non-IID and unbalanced data distribution that makes the training process harder. The difficulty of training Non-IID data is the accuracy reduction. \citet{zhao2018federated} explains that due to the Non-IIDness, the fact of accuracy reduction can be understood in terms of weight divergence, resulting in nonnegligible deviation from correct updates of weights at the stage of averaging. The author also proposes a data-sharing strategy by creating a small globally-shared subset of data. This strategy can effectively improve the accuracy, and for privacy safety, the shared-data can be extracted with distillation\citep{wang2018dataset}, or generated by generative adversarial network (GAN)\citep{chen2019data}. Many theoretical works are also made for FedAvg focusing on convergence analysis and relaxing the assumptions in the Non-IID setting\citep{li2019convergence,li2019communication,lian2017can}. However, all these works focus on training a single global model.
\paragraph{Model Heterogeneity} 
\citet{smith2017federated} introduce a MOCHA framework for multi-task federated learning, dealing with high communication cost, stragglers, and fault tolerance. \citet{khodak2019adaptive} presents an Average Regret-Upper-Bound Analysis (ARUBA) theoretical framework for analyzing gradient-based meta-learning. These frameworks allows separate models training but the architectures of model are still controlled by central server. \citet{li2019fedmd} proposes a decentralized framework based on knowledge distillation, which enables federated learning for independently designed models. However, this method requires a public dataset but have no global model for subsequent use. It also does not support new participation, since new participants may wreck established models.
\paragraph{Objective Heterogeneity}
The objective of the traditional Federated Learning is to train a global model that can be used for all the clients. However, in the personalization situation, \citet{yu2020salvaging} shows that some participants may not benefit from the global model when the global model is less accurate than the local model. For some clients whose local dataset is small, the global model will be overfitted to these local data that influence its personalization ability. \citet{jiang2019improving} point out that optimizing only for the global accuracy will make the model harder to personalize. Therefore, Jiang proposed three following objects for the personalization of Federated learning: 1) developing improved personalized models that benefit a large majority of clients; 2) developing an accurate global model that benefits clients who have limited private data for personalization; 3) attaining fast model convergence in a small number of training rounds. \citet{liu2020federated} proposes an federated learning framework to obtain various types of image representations from different tasks and merge useful features from different vision-and-language grounding problems.

\section{Preliminaries}

\subsection{Typical Federated Learning Setup}
The objective of typical federated learning (FedAvg) is to train a single shared model over decentralized data, by minimizing the global objective function $\min f(w)$ in the distributed manner, where the whole dataset is the union of each decentralized data, and the loss function is over all private data $f(w) = \frac{1}{n} \sum_{i=1}^{n} f_{i}(w)$. Given private data, denoted as  $(\mathcal{X}_k, \mathcal{Y}_k)$ generated by distinct distribution $ \mathcal{P}_k(x, y)$ from $K$ clients, federated learning on each client starts with copying the weight vector $w^k \in \mathbb{R}^d$ from the global model. Each clients then conducts local update that optimizing the local objective by gradient decent method for several epochs:
\begin{equation}
\begin{array}{l}
	F_k(w^k)=\frac{1}{n_k} \sum_{i \in \mathcal{P}_{k}} f_{i}(w^k), \\ 
	w^{k} \leftarrow w^{k}-\eta \nabla F_{k}\left(w^{k}\right),
\end{array}
\end{equation}
where $F_k(w^k)$ is the loss function of the $k$-th client, $n_k$ is the number of local samples, $\eta$ is the learning rate, and $\nabla F_{k}\left(w^{k}\right)\in \mathbb{R}^d$ is the gradient of $F_{k}\left(w^{k}\right)$. Note that the expectation $\mathbb{E}_{\mathcal{P}_{k}}\left[F_{k}(w)\right]=f(w)$ may not hold because $\mathcal{P}_k\not= \mathcal{P}_{joint}$ in the Non-IID setting. After a period of local updates, clients transmit local model weights $w^{k}$ to the parameter server, who then aggregates these weights by weighted averaging:
\begin{equation}
	w^{global} \leftarrow \sum_{k=1}^{K} \frac{n_{k}}{n} w^{k},
\end{equation}
where $w^{global}$ is the weights of the global model, and $n$ is the number of samples over all clients. Repeat the whole training process until the global model gets convergence, and the  shared global model can learn from collaboratively training without sharing private local data. However, the training of local model is directly on the copy of the global model, so the typical federated learning will face the problems previously described. Training distinct models for clients is a natural way to solve these problems.
\subsection{Knowledge Distillation}
Knowledge distillation \citep{hinton2015distilling} is knowledge distillation is the process of transferring dark knowledge from a powerful large teacher model to a lighter easier-to-deploy student model, without significant loss in performance. 
The loss function of student model can be simplified as follows:
\begin{equation}
\begin{array}{c}
L_{student}=L_{CE}+D_{KL}\left({p}_{teacher} \| {p}_{student}\right),\\
{p}_{teacher}=\frac{\exp \left(z / T\right)}{\sum_i \exp \left(z_i / T\right)}
\end{array}
\label{origindml}
\end{equation}
where $L_{CE}$ and $D_{KL}$ are the Cross Entropy and Kullback Leibler (KL) Divergence, ${p}_{teacher}$ and ${p}_{student}$ are the predicts of teacher and student model, $T$ is a hyper-parameter means temperature, and $z$ is the logits of teacher model. This method can improve the performance of student model because the prediction of teacher model can give more helpful information (soft targets) than one-hot label (hard targets), as a regularizer.

We will introduce knowledge distillation into federated learning at the stage of local update, and it is for two reasons: 1) FL can be regarded as a processing of transfer learning among global and local models; 2) The two models that transfer knowledge can have different architectures. However, a well-trained teacher model is not exist in federated learning, thus we adopt deep mutual learning (DML) \citep{zhang2018deep} as our process of local update, which is a deep learning strategy derived from knowledge distillation. Different from the one-way knowledge transfer that from a well-trained teacher to an untrained student, DML is not the teacher-to-student pattern, but the two models can learn from each other throughout the training process. The loss function of two models as follows:

\begin{equation}
\begin{array}{l}
L_{w^{1}}=L_{C_{1}}+D_{K L}\left({p}_{2} \| {p}_{1}\right), \\
L_{w^{2}}=L_{C_{2}}+D_{K L}\left({p}_{1} \| {p}_{2}\right),
\end{array}
\label{origindml}
\end{equation}
where $p_1$ and $p_2$ are the predicts of two networks. The objective of the two models is training themselves over dataset meanwhile achieving predicts consensus (distillation). With DML, the two models can get better performance than independent training. Note that the two models can also have different architecture, as well as the direction of knowledge transfer is two-way. Thus, DML can help train distinct models during local update in federated learning.

\section{Methodology}
\subsection{Rethinking Federated Learning}
In order to deal with the three heterogeneities (DOM), we are trying to rethink these two questions in federated learning?
\paragraph{What is the product of FL?} In typical FL setting, the objective is training a single out-of-the-box (OOTB) model by and for all clients, to fits a joint distribution $\mathcal{P}_{joint}(x,y)$. For this single objective, the Non-IIDness of data (DH) brings difficulties to training. However, in the sense of OH, server and clients have different objectives, where server aims to a generalized model that fits $\mathcal{P}_{joint}(x,y)$ whereas clients aim to a personalized model that fits $\mathcal{P}_{k}(x,y)$. Therefore, \emph{the Non-IIDness of data (DH) is harmful for server but is beneficial for clients}, if it is able to train a personalized model in FL. Thus, the Non-IIDness of data is no longer a bug but a feature that clients can be personally served better.
\paragraph{What to shared in FL?} Inspired by \citep{li2020knowledge, gao2019hhhfl,liu2020federated}, we find that the model shared in FL does not have to be a complete model, e.g. an end-to-end (E2E) model. The model trained by FL could be split into two parts: 1) a partial model shared globally and 2) a partial model owned by clients locally, depending on what clients want to learn and share in FL. The shared objective could be an encoder for learning representations, a decoder \citep{gao2019hhhfl, liang2020think} for classification or an integrated module \citep{liu2020federated} for multi-task learning. However, the different local objectives (OH) of clients would give rise to the requirement of MH (e.g. clients may have similar but different tasks like VQA and Image caption \citep{liu2020federated}).

\subsection{Federated Mutual Learning}\label{fmlsec}

With the three heterogeneities (DOM), it is necessary to enable FL to train generalized and personalized model mutually with different architectures. In this section, we introduce a more flexible federated learning method, named Federated Mutual Learning (FML), that deal with the three heterogeneities. 

\begin{figure*}[t!]
	\centering
	\includegraphics[width=\textwidth]{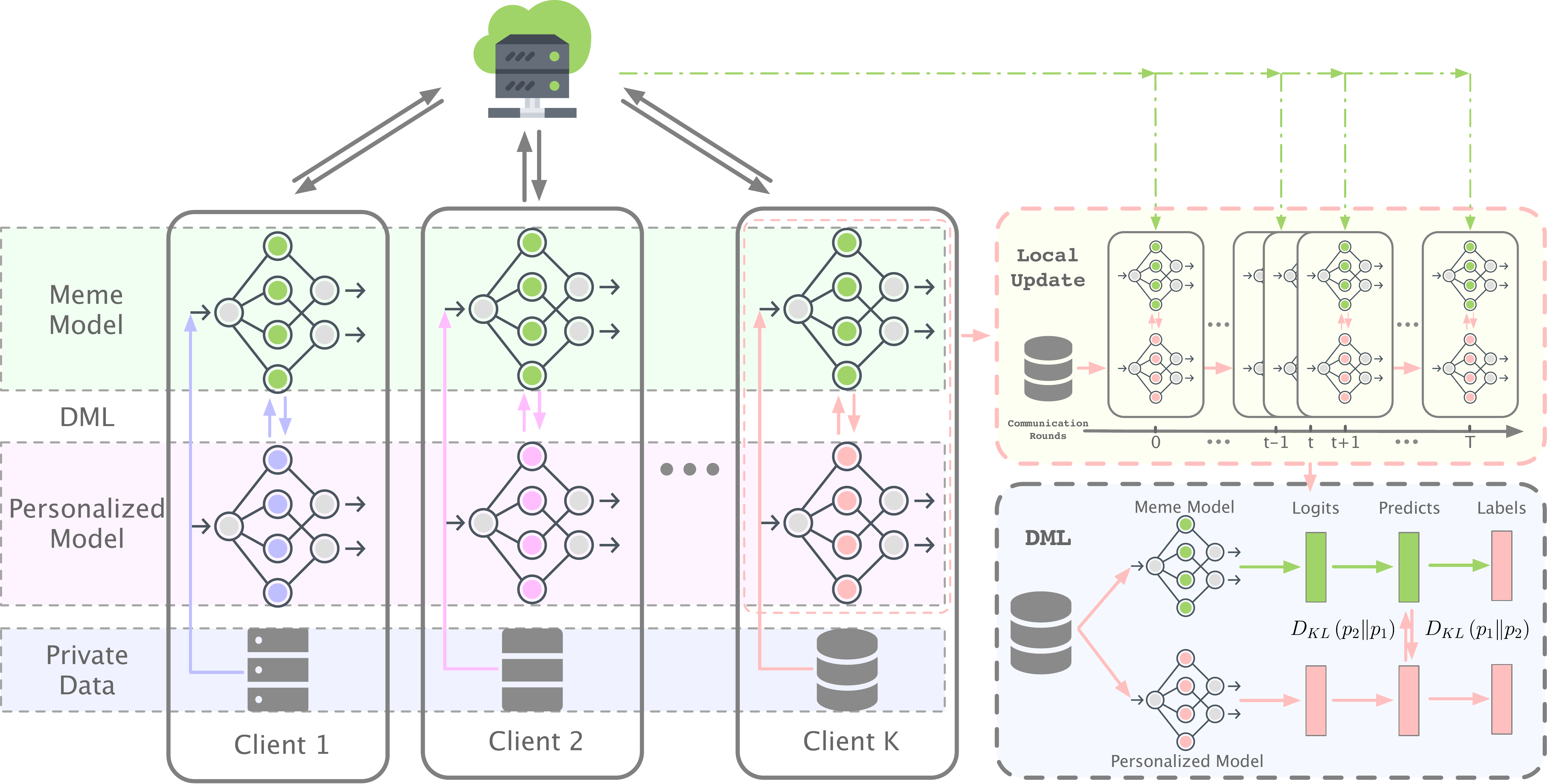}
	\caption{a) Each client in FML trains two models over private data during local update: the meme model and the personalized model; b) At each communication round, clients fork the new generation of global model as its meme model but the personalized model is trained privately and continuously; c) During each local update, the two models in clients conduct DML for several epochs, learning mutually.}\label{fml}
\end{figure*}

Federated learning can be regarded as a method where the knowledge of data is learned and transferred between global model and local models. Thus, we introduce a knowledge distillation method, deep mutual learning (DML), as the approach of local update of clients, so that clients can train a personalized model for its own data and task. For each client in FML, there are two models inside (Figure~\ref{fml}): One is the meme model that is the medium of knowledge transfer between global and local models; another is the personalized model that clients designed for their own data and task. Thus, FML allows clients train local model mutually with global model rather than directly on it. 

During the process of training, FML starts with an initial global model, which is controlled by the central server. At the meantime, all clients start with an initial personalized model, which is customized by each client or use the same model provided by server (for convenience). Then, all clients fork the global model as its meme model and conduct local update. Note that if the global model is not a complete model, the meme model is built by splicing the forked global model with an adaptor layer (e.g. if the global model is just the convolution layers of CNN model, the meme model is the convolution layers and a FC layers). Next, instead of directly training on the copy of the global model, the local update of each client is conducting DML between the meme model and personalized model for several epochs. We rewrite the loss function of the two models as follows:
\begin{equation}
\begin{array}{c}
L_{local}=\alpha L_{C_{local}}+(1-\alpha)D_{K L}\left({p}_{meme} \| {p}_{local}\right),\\
L_{meme}=\beta L_{C_{meme}}+(1-\beta)D_{K L}\left({p}_{local} \| {p}_{meme}\right), 
\end{array}
\end{equation}
where $\alpha$ and $\beta$ are the hyper-parameters that control the proportion of knowledge that from data or from the other model. The direction of knowledge transfer between meme and local model is two-way, that meme model will transfer global knowledge to personalized model and will get feedback from it, and they are both trained over private data. Finally, each clients push its trained meme model to server, and the server merge these meme models by averaging into the new generation of global model. Repeat the whole process until convergence, and the whole process shown in Algorithm 1.

Where $global$ is the global model, $meme$ is the meme model, $local$ is the personalized model, $(\mathcal{X}, \mathcal{Y})$ is the private data, the subscript $t$ denotes the $t$-th communication round, the superscript $k$ denotes the $k$-th client, $e$ denotes the $e$-th epoch of local update, the upper case letters $T, K, E$ are the maximum value of $t, k, e$. Note that FML would degrade into typical FedAvg if $\beta=1$, and the weighted average item $\frac{n_k}{n}$ in FedAvg is abandoned in our approach, which discussed in Section \ref{discuss}. From the perspective of server, the global model is learned by FedAvg with meme models of clients, which is the generalized model fitting the joint distribution $\mathcal{P}_{joint}(x, y)$ over all data. From the perspective of clients, the personalized models conduct continuously training over private data, and continuously distilling knowledge from meme models at each communication round, shown in Figure \ref{fml}. Throughout the whole process, the personalized models will not be replaced and never leave the clients, fitting the personalized distribution $\mathcal{P}_{k}(x, y)$ over private data.  

\begin{algorithm}[t]
\renewcommand{\algorithmicrequire}{\textbf{Server executes:}}
\renewcommand{\algorithmicensure}{\textbf{ClientUpdate:}}
    \caption{\emph{Federated Mutual Learning (FML)}:}
\begin{algorithmic}[1]
\Require \\
initialize the global model $global_0$
\For{each round $t=1,2,\dots, T$}
\For {each client $k$ \textbf{in parallel}}
\State $meme_{t+1}^k \gets$ ClientUpdate($meme^k_t$)
\EndFor
\State Merge: $global_{t+1}\gets\frac{1}{K}\sum^K_{k=1}meme_{t+1}^k$
\EndFor
\end{algorithmic}
\begin{algorithmic}[1]
\Ensure\\
initialize the local personalized model $local^k_0$
\State Fork: $meme^k_t \gets global_t$
\For {each epoch $e=1,2,\dots, E$}
\State Conduct DML between $meme^k_t \& local^k_t$ over private data $(\mathcal{X}_k, \mathcal{Y}_k)$
\EndFor
\end{algorithmic}
\end{algorithm}
\section{Experiments}
In this section we validate the performance of FML over three commonly used image classification datasets in IID and Non-IID settings implemented with PyTorch. The experiments are designed in two parts: 1) we validate the performance of FML in typical FL settings; 2) we validate FML in the settings of DOM. 
\subsection{Experiment settings} 
\paragraph{Datasets}
We use MNIST and CIFAR10/100 as the datasets of federated learning. MNIST \citep{lecun1998gradient} is a 10-classification handwriting digit dataset, with the number of 0 to 9. The dataset consists of 60000 training images and 10000 testing images, with 6000 and 1,000 images (28x28 pixels) per digit, respectively. CIFAR10/100 \citep{krizhevsky2009learning} are 10/100 classification datasets respectively. CIFAR10 consists of 50000 training images and 10000 test images in 10 classes, with 5000 and 1000 images per class. CIFAR100 has the same total number of images as CIFAR10, but it has 100 classes. CIFAR100 has 500 training images and 100 testing images per class. All images of CIFAR10/100 are 3-channel color images (32x32). These three datasets are widely used in federated learning.

\paragraph{Federated settings}
We configure our experiments with a simulated cross-silo federated learning environment. There are five clients ($K=5$) under the orchestration of a central server. The total communication rounds $T=200$, and the number of local epochs $E=5$. We divide the dataset into five parts for five clients in IID and Non-IID setting. $1/5$ training data is equally distributed to each client as its private training data, and $1/5$ test data is also distributed as its private validate data in the same setting (e.g. CIFAR10 has totally 50000 training images and 10000 test images, then each client gets 10000 private training data and 2000 private validate data in IID setting without replacement). The overall test set (e.g. 10000 test images of CIFAR10) is also used for testing global model. For IID setting, each client gets a shuffled data that $\mathcal{P}_k(x, y)= \mathcal{P}_{joint}(x, y)$. For Non-IID setting, we consider three levels of difficulty: we divide the dataset by label into $K \cdot p$ shards with the size of $\frac{n}{K\cdot p}$ (in our experiments $K=5$ and $n=50000$), and assign $p$ shards to each client ($p=\left \{6,4,2\right \}$ for MNIST and CIFAR10, $p=\left \{60,40,20\right \}$ for CIFAR100, which means each client have at most $p$ classes of data). We denote them as Non-IID (1, 2, 3), where Non-IID (3) is an extreme setting that there is no overlap of classes. 

\paragraph{Training settings}
We use four types of model in the experiments: multi-layer perceptron (MLP)  \citep{mcmahan2016communication}, LeNet5 \citep{lecun1990handwritten}, a convolutional neural networks (CNN1) with two 3x3 convolution layers (the first with 6 channels, the second with 16 channals, each followed with 2x2 max pooling and ReLu activation) and two FC layers, and a convolutional neural networks (CNN2) with three 3x3 convolution layers (each with 128 channels followed with 2x2 max pooling and ReLu activation) and one FC layer. MLP/LeNet5 are used for MNIST dataset and CNN1/CNN2 are used for CIFAR10/100 datasets. The optimizer we choose is the SGD algorithm, with $momentum=0.9$, $weight\_decay=5\times 10^{-4}$ and $batchsize=128$.

\subsection{FML in typical FL settings}
\begin{figure*}[!t]
	\centering
	\includegraphics[width=\textwidth]{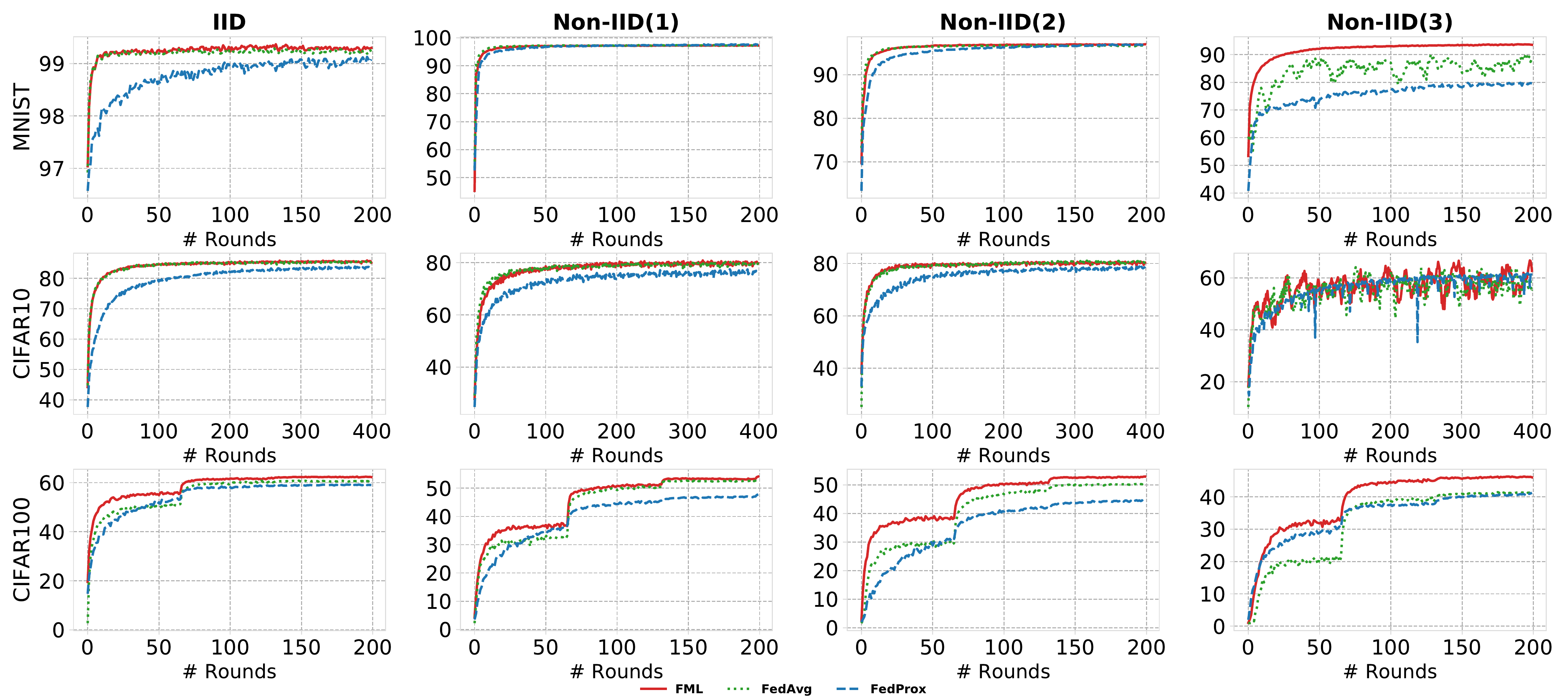}
	\caption{FML results in better improvements than FedAvg and FedProx in four data settings. We simulate different levels of data heterogeneity (it becomes more difficult from left to right). Due to the DML, the $D_{KL}$ loss item is a strong regularizer for training. In the Non-IID setting, FML performs better with a steady trajectory, than FedProx and FedAvg that with severe oscillation. According to \cite{zhang2018deep}, FML can find a more steady (robust) minimum.}\label{fmltfl}
	\includegraphics[width=\textwidth]{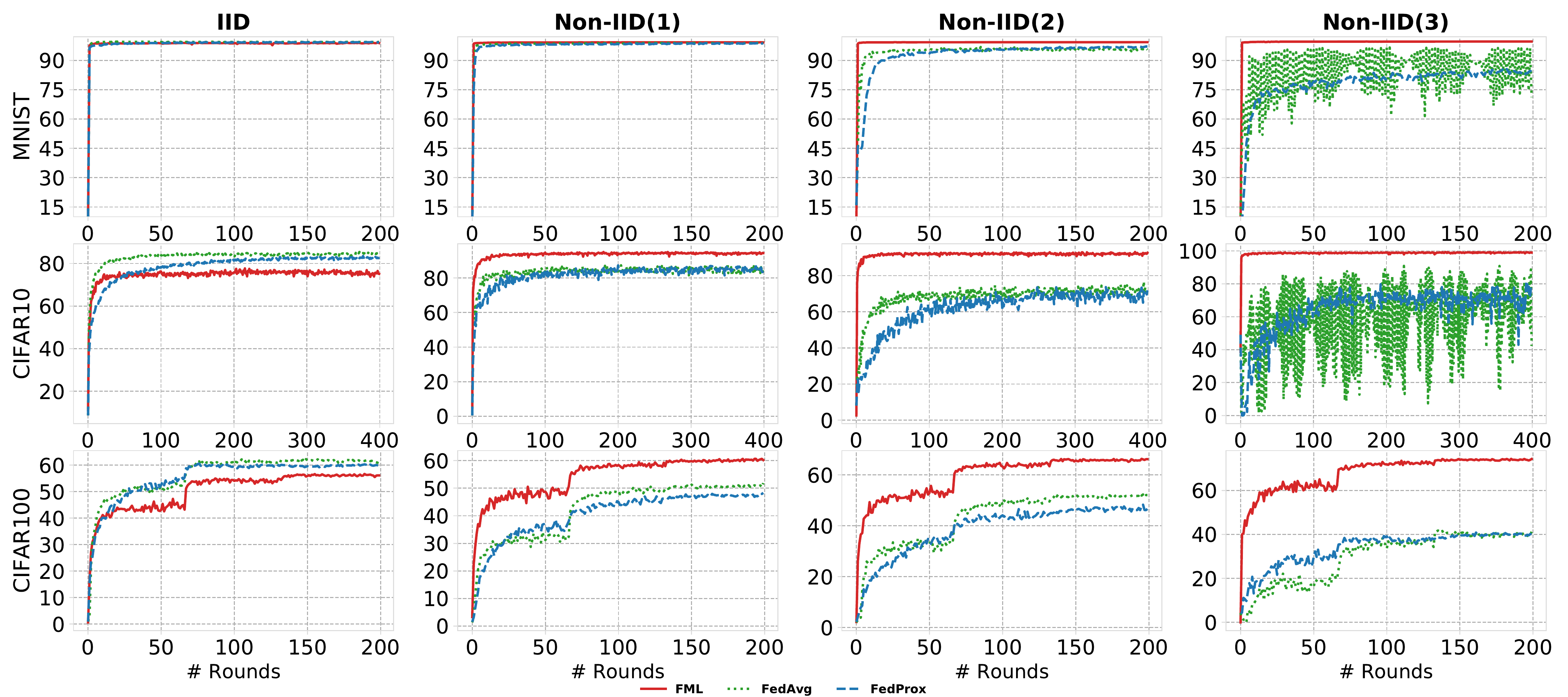}
	\caption{The global model is trained as generalized model, who has poor performance on private data in Non-IID settings. Comparing the three curves, FedAvg oscillate more severely due to the increasing level of data heterogeneity of data, especially in Non-IID(3); Fedprox adds a proximal term to alleviating the oscillation but fails to achieve high accuracy; FML rises rapidly and stables at a high level, performing better both in stability and accuracy.}\label{fmldh}
\end{figure*}

We first study FML in typical FL settings, where all clients aim to train a shared global model. In the experiment, we initialize all models (the global model, meme model and personalized model of each client) with the same architecture. We perform FedAvg \citep{mcmahan2016communication} and FedProx \citep{li2018federated} as two baselines and compare them with FML in both IID and Non-IID manners. We test four types of model (MLP, LeNet5, CNN1, CNN2) over three datasets (MNIST, CIFAR10, CIFAR100) in four data settings (IID, Non-IID(1, 2, 3)), and the accuracy of global model are exhibited in Table \ref{global}. Comparing IID with Non-IID (1, 2, 3), the accuracy of global model is decreasing with the increasing level of difficulty of data setting from the top to the bottom of the Table. Comparing FML with baselines, FML shows better results than that of FedAvg and FedProx in most of settings.  Figure \ref{fmltfl} shows the training process of the experiments.

\begin{table*}[t]
  \centering
  \setlength{\tabcolsep}{4mm}{
  \begin{tabular}{c|c|cc|cc|cc}
    \toprule
    \multirow{2}*{Settings}  & \multirow{2}*{Methods}   & \multicolumn{2}{|c|} {MNIST}   & \multicolumn{2}{|c|}{CIFAR10}  & \multicolumn{2}{c}{CIFAR100}   \\
    \cmidrule(r){3-8}
     &        & MLP & LeNet5 & CNN1  &CNN2 &CNN1     & CNN2 \\
    \midrule
    \multirow{3}*{IID}&FedAvg &98.44&99.29 &85.90 &87.49&56.11&60.88\\
    &FedProx    & 98.14&99.13&83.91&86.15&32.41&59.23  \\

    &FML(ours)    & \textbf{98.49}& \textbf{99.37}& \textbf{85.93}&87.41&\textbf{57.11}&\textbf{62.50} \\
    \midrule
    \multirow{3}*{Non-IID (1)}&FedAvg & 97.40 & 98.92 &80.41&82.64  &53.77& 57.76  \\
    &FedProx    & 97.35 & 98.75 & 77.46& 80.88& 47.83& 55.60  \\

    &FML(ours)    & \textbf{97.70} & \textbf{99.07} &\textbf{80.86}&\textbf{82.69}  & \textbf{54.21}&\textbf{59.77}   \\
    \midrule
    \multirow{3}*{Non-IID (2)}&FedAvg  & 96.84&  98.67&78.85& 81.17&50.86&56.82\\
    &FedProx    & 96.98 &98.50&76.53&78.87&45.46&55.34\\

    &FML(ours)    & \textbf{97.00} &\textbf{98.71}&78.64&80.85&\textbf{52.92}&55.93\\
    \midrule
 \multirow{3}*{Non-IID (3)}&FedAvg & 90.46 &96.45&63.22&64.12&41.48&50.36\\
    &FedProx    & 80.03&87.55&58.07&62.01&41.29&49.51\\
    &FML(ours)    & \textbf{93.77} & \textbf{96.70}&62.42&\textbf{66.75}&\textbf{46.30}&\textbf{51.86}\\
    \bottomrule
  \end{tabular}
  }
  \caption{Top-1 accuracies (\%) of global model in typical FL settings}\label{global}
\end{table*}
\begin{figure}[!h]
	\centering
	\includegraphics[width=0.5\textwidth]{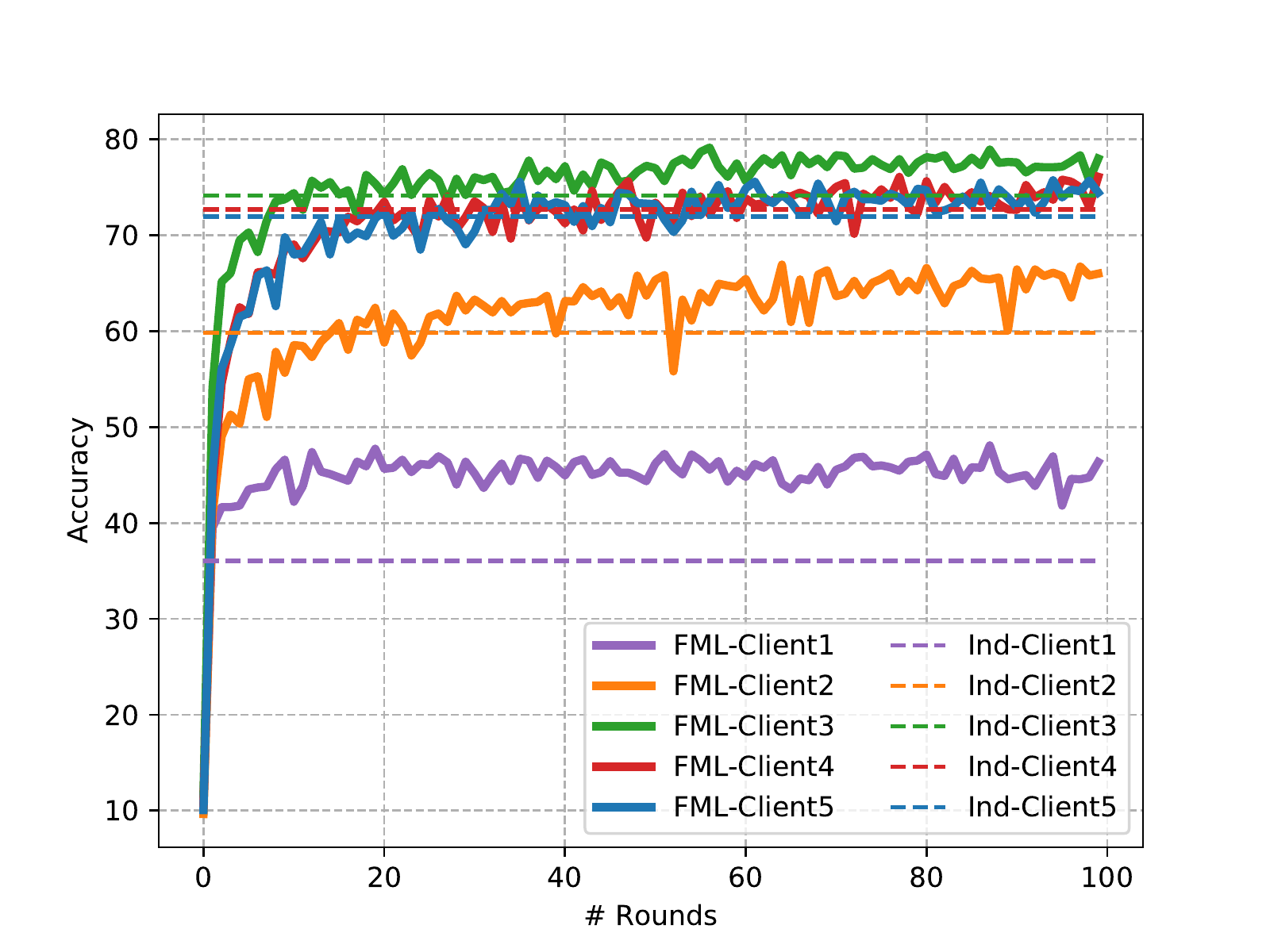}
	\caption{The solid curve is the accuracy of personalized model training with FML, and the dash line is the best accuracy of personalized model with independently training, both over private validate set. The results shows that FML can benefit all clients with different models by a shared model.}\label{fmlmh}
\end{figure}
\begin{figure}[!h]
	\centering
	\includegraphics[width=0.5\textwidth]{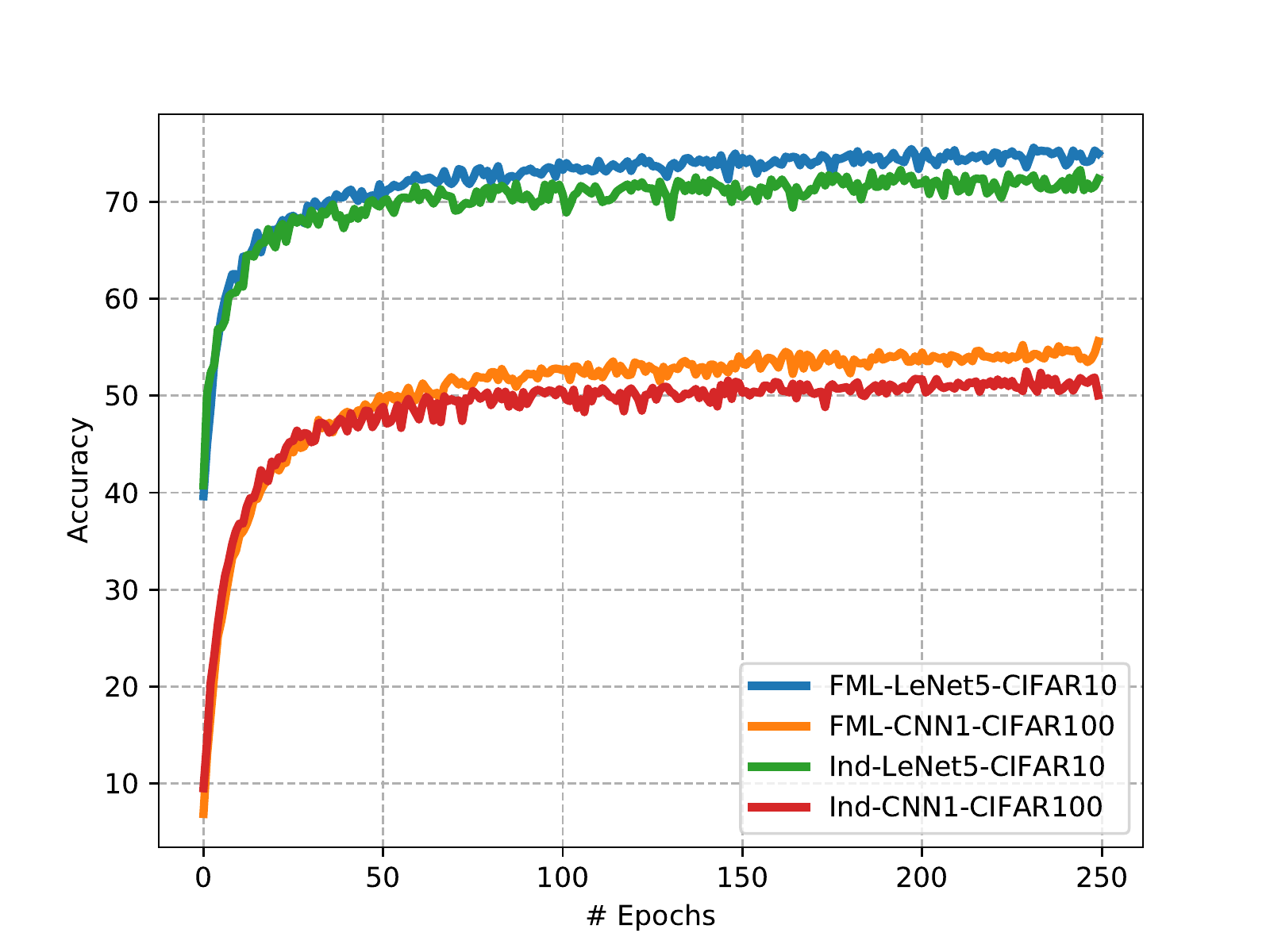}
	\caption{The green and red curves refer to LeNet5 and CNN1 trained independently over CIFAR10 and CIFAR100, respectively. The blue and orange curves refer to the two models trained by FML. The results shows that FML can benefit all clients with different tasks by a shared representation.}\label{fmloh}
\end{figure}
\subsection{FML in DOM}
\paragraph{Data Heterogeneity} Due to the Non-IIDness of data, the global shared model may perform worse than local models that solely trained on their private data. Thus, the way we deal with data heterogeneity is by training a personalized model for each client, since it allows them fits their own distribution $\mathcal{P}_{k}(x, y)$, rather than a single generalized model in typical FL setting. 
We test the performance of the product of FL (the personalized model in FML and the global model in FedAvg and FedProx) on private validate set, shown in Figure \ref{fmldh}.
\paragraph{Model Heterogeneity} We deal with the model heterogeneity by introducing knowledge distillation method, so that each client can design their personalized model on demand. Here we setup different models for different clients: We set LeNet5 as the global model, and 1xMLP, 1xLeNet5, 1xCNN1, 2xCNN2 for five clients. The five models are trained over CIFAR10 in IID setting. The results of this experiment are shown in Figure \ref{fmloh}. 
\paragraph{Objective Heterogeneity} Here we study the multi-task FML, where clients may run different tasks. We initialize two clients that train 10-/100-classification (CIFAR10/100) tasks with LeNet5 and CNN1, respectively. The global model is only the convolution layers of CNN2. Note that the global model is not a complete end-to-end model, thus an Adaptor layer (an FC layer) have to be spliced on it, for adapting 10-/100-classification task of each client. Figure \ref{fmloh} shows the results.

\section{Discussion}\label{discuss}
\paragraph{Catfish Effect}
Since the FML can deal with model heterogeneity, various clients allows training models with different dimensions and architecture. The capabilities of models of various clients may be different. In our experiment, we observed a natural or social phenomena that never happens
in FedAvg in some case, catfish effect, that models with low capabilities (sardines) can be improved by a model with high capability (catfish), compared to FML with only sardines. 
On the contrary, if there exists a badly-trained model in FML, the overall performance of other clients have little effect. 
This feature may derive some research about adversarial training in federated learning in the future.
\paragraph{Dynamic Alphabeta}
In our experiments, the proportions of cross entropy loss and KL loss of local ($\alpha$) and meme model ($\beta$) are fixed. However, we find it significantly important for training local and global model. Dynamic alphabeta at different stage of training can improve both global and local performance. From our experience, the improvement of local model attribute to well-trained global model at later stage, whereas the improvement of global model attribute to well-trained local models at early stage. Thus, a larger $\alpha$ in early stage and a larger $\beta$ in later stage is preferred.
\paragraph{Privacy and Fairness}
In this paper, we introduce the model privacy. FML allows customized models, which is also the  private property of individuals, so that local customized model should be protected from being stolen. In addition, the average item $\frac{n_k}{n}$ in Section \ref{fmlsec} is abandoned in consideration of privacy and fairness. One the one hand, the number of samples $n_k$ on each client should not be exposed to central server, since it might be an auxiliary for stealing privacy by attacker; on the other hand, the different $n_k$ leads to fairness problem, that client with large mount of samples will take a big part in model training, which is not appropriate in some applications. Instead, we abandon this item and consider each client as equal rather than each sample.


\section*{Ethical Impact}
Federated learning is not only a technical standard, but also a ``win-win'' business model. Federated learning, as the underlying technology for Al's development, can drive cross-disciplinary enterprise-level data cooperation and help enterprises participate in the globalization. The cross-silo federated learning can be benefited from FML, since clients may have rather sufficient data and need to design customized local model. On the contrary, the cross-device federated learning might not suitable for FML, since clients may not have the requirement of customized models.



\begin{thebibliography}{27}

\bibitem[McMahan et~al.(2016)McMahan, Moore, Ramage, Hampson,
  et~al.]{mcmahan2016communication}
H~Brendan McMahan, Eider Moore, Daniel Ramage, Seth Hampson, et~al.
\newblock Communication-efficient learning of deep networks from decentralized
  data.
\newblock \emph{arXiv preprint arXiv:1602.05629}, 2016.

\bibitem[Kairouz et~al.(2019)Kairouz, McMahan, Avent, Bellet, Bennis, Bhagoji,
  Bonawitz, Charles, Cormode, Cummings, et~al.]{kairouz2019advances}
Peter Kairouz, H~Brendan McMahan, Brendan Avent, Aur{\'e}lien Bellet, Mehdi
  Bennis, Arjun~Nitin Bhagoji, Keith Bonawitz, Zachary Charles, Graham Cormode,
  Rachel Cummings, et~al.
\newblock Advances and open problems in federated learning.
\newblock \emph{arXiv preprint arXiv:1912.04977}, 2019.

\bibitem[Lim et~al.(2020)Lim, Luong, Hoang, Jiao, Liang, Yang, Niyato, and
  Miao]{lim2020federated}
Wei Yang~Bryan Lim, Nguyen~Cong Luong, Dinh~Thai Hoang, Yutao Jiao, Ying-Chang
  Liang, Qiang Yang, Dusit Niyato, and Chunyan Miao.
\newblock Federated learning in mobile edge networks: A comprehensive survey.
\newblock \emph{IEEE Communications Surveys \& Tutorials}, 2020.

\bibitem[Yang et~al.(2019)Yang, Liu, Cheng, Kang, Chen, and
  Yu]{yang2019federated}
Qiang Yang, Yang Liu, Yong Cheng, Yan Kang, Tianjian Chen, and Han Yu.
\newblock Federated learning.
\newblock \emph{Synthesis Lectures on Artificial Intelligence and Machine
  Learning}, 13\penalty0 (3):\penalty0 1--207, 2019.

\bibitem[Zhao et~al.(2018)Zhao, Li, Lai, Suda, Civin, and
  Chandra]{zhao2018federated}
Yue Zhao, Meng Li, Liangzhen Lai, Naveen Suda, Damon Civin, and Vikas Chandra.
\newblock Federated learning with non-iid data.
\newblock \emph{arXiv preprint arXiv:1806.00582}, 2018.

\bibitem[Wu et~al.(2019)Wu, Dai, Zhang, Wang, Sun, Wu, Tian, Vajda, Jia, and
  Keutzer]{wu2019fbnet}
Bichen Wu, Xiaoliang Dai, Peizhao Zhang, Yanghan Wang, Fei Sun, Yiming Wu,
  Yuandong Tian, Peter Vajda, Yangqing Jia, and Kurt Keutzer.
\newblock Fbnet: Hardware-aware efficient convnet design via differentiable
  neural architecture search.
\newblock In \emph{Proceedings of the IEEE Conference on Computer Vision and
  Pattern Recognition}, pages 10734--10742, 2019.

\bibitem[He et~al.(2020)He, Annavaram, and Avestimehr]{he2020fednas}
Chaoyang He, Murali Annavaram, and Salman Avestimehr.
\newblock Fednas: Federated deep learning via neural architecture search.
\newblock \emph{arXiv preprint arXiv:2004.08546}, 2020.

\bibitem[Liang et~al.(2020)Liang, Liu, Ziyin, Salakhutdinov, and
  Morency]{liang2020think}
Paul~Pu Liang, Terrance Liu, Liu Ziyin, Ruslan Salakhutdinov, and
  Louis-Philippe Morency.
\newblock Think locally, act globally: Federated learning with local and global
  representations.
\newblock \emph{arXiv preprint arXiv:2001.01523}, 2020.

\bibitem[Gao et~al.(2019)Gao, Ju, Wei, Liu, Chen, and Yang]{gao2019hhhfl}
Dashan Gao, Ce~Ju, Xiguang Wei, Yang Liu, Tianjian Chen, and Qiang Yang.
\newblock Hhhfl: Hierarchical heterogeneous horizontal federated learning for
  electroencephalography.
\newblock \emph{arXiv preprint arXiv:1909.05784}, 2019.

\bibitem[Smith et~al.(2017)Smith, Chiang, Sanjabi, and
  Talwalkar]{smith2017federated}
Virginia Smith, Chao-Kai Chiang, Maziar Sanjabi, and Ameet~S Talwalkar.
\newblock Federated multi-task learning.
\newblock In \emph{Advances in Neural Information Processing Systems}, pages
  4424--4434, 2017.

\bibitem[Wang et~al.(2018)Wang, Zhu, Torralba, and Efros]{wang2018dataset}
Tongzhou Wang, Jun-Yan Zhu, Antonio Torralba, and Alexei~A Efros.
\newblock Dataset distillation.
\newblock \emph{arXiv preprint arXiv:1811.10959}, 2018.

\bibitem[Chen et~al.(2019)Chen, Wang, Xu, Yang, Liu, Shi, Xu, Xu, and
  Tian]{chen2019data}
Hanting Chen, Yunhe Wang, Chang Xu, Zhaohui Yang, Chuanjian Liu, Boxin Shi,
  Chunjing Xu, Chao Xu, and Qi~Tian.
\newblock Data-free learning of student networks.
\newblock In \emph{Proceedings of the IEEE International Conference on Computer
  Vision}, pages 3514--3522, 2019.

\bibitem[Li et~al.(2019{\natexlab{a}})Li, Huang, Yang, Wang, and
  Zhang]{li2019convergence}
Xiang Li, Kaixuan Huang, Wenhao Yang, Shusen Wang, and Zhihua Zhang.
\newblock On the convergence of fedavg on non-iid data.
\newblock \emph{arXiv preprint arXiv:1907.02189}, 2019{\natexlab{a}}.

\bibitem[Li et~al.(2019{\natexlab{b}})Li, Yang, Wang, and
  Zhang]{li2019communication}
Xiang Li, Wenhao Yang, Shusen Wang, and Zhihua Zhang.
\newblock Communication efficient decentralized training with multiple local
  updates.
\newblock \emph{arXiv preprint arXiv:1910.09126}, 2019{\natexlab{b}}.

\bibitem[Lian et~al.(2017)Lian, Zhang, Zhang, Hsieh, Zhang, and
  Liu]{lian2017can}
Xiangru Lian, Ce~Zhang, Huan Zhang, Cho-Jui Hsieh, Wei Zhang, and Ji~Liu.
\newblock Can decentralized algorithms outperform centralized algorithms? a
  case study for decentralized parallel stochastic gradient descent.
\newblock In \emph{Advances in Neural Information Processing Systems}, pages
  5330--5340, 2017.

\bibitem[Khodak et~al.(2019)Khodak, Balcan, and Talwalkar]{khodak2019adaptive}
Mikhail Khodak, Maria-Florina~F Balcan, and Ameet~S Talwalkar.
\newblock Adaptive gradient-based meta-learning methods.
\newblock In \emph{Advances in Neural Information Processing Systems}, pages
  5915--5926, 2019.

\bibitem[Li and Wang(2019)]{li2019fedmd}
Daliang Li and Junpu Wang.
\newblock Fedmd: Heterogenous federated learning via model distillation.
\newblock \emph{arXiv preprint arXiv:1910.03581}, 2019.

\bibitem[Yu et~al.(2020)Yu, Bagdasaryan, and Shmatikov]{yu2020salvaging}
Tao Yu, Eugene Bagdasaryan, and Vitaly Shmatikov.
\newblock Salvaging federated learning by local adaptation.
\newblock \emph{arXiv preprint arXiv:2002.04758}, 2020.

\bibitem[Jiang et~al.(2019)Jiang, Kone{\v{c}}n{\`y}, Rush, and
  Kannan]{jiang2019improving}
Yihan Jiang, Jakub Kone{\v{c}}n{\`y}, Keith Rush, and Sreeram Kannan.
\newblock Improving federated learning personalization via model agnostic meta
  learning.
\newblock \emph{arXiv preprint arXiv:1909.12488}, 2019.

\bibitem[Liu et~al.(2020)Liu, Wu, Ge, Fan, and Zou]{liu2020federated}
Fenglin Liu, Xian Wu, Shen Ge, Wei Fan, and Yuexian Zou.
\newblock Federated learning for vision-and-language grounding problems.
\newblock In \emph{AAAI}, pages 11572--11579, 2020.

\bibitem[Hinton et~al.(2015)Hinton, Vinyals, and Dean]{hinton2015distilling}
Geoffrey Hinton, Oriol Vinyals, and Jeff Dean.
\newblock Distilling the knowledge in a neural network.
\newblock \emph{arXiv preprint arXiv:1503.02531}, 2015.

\bibitem[Zhang et~al.(2018)Zhang, Xiang, Hospedales, and Lu]{zhang2018deep}
Ying Zhang, Tao Xiang, Timothy~M Hospedales, and Huchuan Lu.
\newblock Deep mutual learning.
\newblock In \emph{Proceedings of the IEEE Conference on Computer Vision and
  Pattern Recognition}, pages 4320--4328, 2018.

\bibitem[Li and Bilen(2020)]{li2020knowledge}
Wei-Hong Li and Hakan Bilen.
\newblock Knowledge distillation for multi-task learning.
\newblock \emph{arXiv preprint arXiv:2007.06889}, 2020.

\bibitem[LeCun et~al.(1998)LeCun, Bottou, Bengio, and
  Haffner]{lecun1998gradient}
Yann LeCun, L{\'e}on Bottou, Yoshua Bengio, and Patrick Haffner.
\newblock Gradient-based learning applied to document recognition.
\newblock \emph{Proceedings of the IEEE}, 86\penalty0 (11):\penalty0
  2278--2324, 1998.

\bibitem[Krizhevsky et~al.(2009)Krizhevsky, Hinton,
  et~al.]{krizhevsky2009learning}
Alex Krizhevsky, Geoffrey Hinton, et~al.
\newblock Learning multiple layers of features from tiny images.
\newblock 2009.

\bibitem[LeCun et~al.(1990)LeCun, Boser, Denker, Henderson, Howard, Hubbard,
  and Jackel]{lecun1990handwritten}
Yann LeCun, Bernhard~E Boser, John~S Denker, Donnie Henderson, Richard~E
  Howard, Wayne~E Hubbard, and Lawrence~D Jackel.
\newblock Handwritten digit recognition with a back-propagation network.
\newblock In \emph{Advances in neural information processing systems}, pages
  396--404, 1990.

\bibitem[Li et~al.(2018)Li, Sahu, Zaheer, Sanjabi, Talwalkar, and
  Smith]{li2018federated}
Tian Li, Anit~Kumar Sahu, Manzil Zaheer, Maziar Sanjabi, Ameet Talwalkar, and
  Virginia Smith.
\newblock Federated optimization in heterogeneous networks.
\newblock \emph{arXiv preprint arXiv:1812.06127}, 2018.

\end{thebibliography}

\end{document}